\documentclass[runningheads]{llncs}

\usepackage{eccv}

\usepackage{eccvabbrv}
\usepackage{enumitem}
\usepackage{graphicx}
\usepackage{booktabs}

\usepackage[accsupp]{axessibility}  

\usepackage{hyperref}

\usepackage{orcidlink}

\usepackage[ruled,linesnumbered]{algorithm2e}
\usepackage{amsmath,amssymb}
\usepackage{xcolor}
\usepackage{graphicx}
\usepackage{float}
\usepackage{subcaption}
\usepackage[most]{tcolorbox}
\usepackage{multirow}
\usepackage{bm}

\newcommand{\revision}[1]{#1} 

\newcommand{\method}{\textsc{CMuon}\xspace} %

\begin{document}


\title{CMuon: Accelerating and Stabilizing \\ Diffusion Transformer Training via \\ Chunked Momentum Orthogonalization}

\titlerunning{CMuon}

\author{Chuyan Chen\inst{1} \quad
Peng Sun\inst{2,3}\textsuperscript{*} \quad
Kun Yuan\inst{1}\textsuperscript{*\dag}
}
\authorrunning{C.~Chen et al.}
%
\institute{ \textsuperscript{1}Peking University \quad
\textsuperscript{2}Westlake University \quad
\textsuperscript{3}Zhejiang University \\
\email{chuyanchen@stu.pku.edu.cn} \quad
\email{sunpeng@westlake.edu.cn} \quad
\email{kunyuan@pku.edu.cn}
}

\maketitle

\renewcommand{\thefootnote}{*}
\footnotetext{These authors jointly supervised this work.}
\renewcommand{\thefootnote}{\dag}
\footnotetext{Corresponding author.}
\renewcommand{\thefootnote}{\arabic{footnote}}
\setcounter{footnote}{0}

\begin{abstract}
  Diffusion Transformers (DiTs) have achieved state-of-the-art (SOTA) performance in visual generative modeling, yet their training remains computationally prohibitive. While the recently proposed Momentum Orthogonalization (Muon) optimizer offers a promising alternative to AdamW, its direct application to DiTs yields suboptimal late-stage convergence. In this paper, we identify the root cause of this bottleneck: standard DiT architectures fuse functionally distinct weights (e.g., within AdaLN and QKV layers) into unified tensors for computational efficiency. Applying Muon to these fused tensors inadvertently induces implicit subspace coupling, which distorts update directions and degrades global optimization. To address this, we introduce Chunked Muon (\method), a simple yet highly effective strategy that partitions these matrices into independent sub-components prior to orthogonalization. Extensive experiments demonstrate that a 675M-parameter DiT trained with CMuon achieves a FID of 1.18 on ImageNet 256 in just 200 epochs. This represents more than a 2x training speedup over AdamW, while effectively overcoming the late-stage convergence plateaus of vanilla Muon.
\end{abstract}

\section{Introduction}
\label{sec:intro}

Training large-scale neural networks with AdamW \cite{loshchilov2017decoupled, kingma2014adam} has long been the standard practice. However, the recent introduction of the Muon optimizer \cite{jordan2024muon} is beginning to shift this paradigm. By orthogonalizing momentum matrices, Muon accelerates convergence and improves memory efficiency compared to AdamW. It has been rapidly adopted within the large language model (LLM) and deep learning communities \cite{team2025kimi, liu2025muon, wu2025hunyuanvideo}, demonstrating significant speedups in large-scale pre-training \cite{team2025kimi,liu2025muon}. Building on this success, several advanced variants—such as Scion \cite{pethick2025training}, Gluon \cite{riabinin2025gluon}, and PolarExpress \cite{amsel2025polar}—have further generalized its capabilities. Consequently, optimization based on Linear Minimization Oracles (LMOs) is rapidly emerging as a powerful new standard for training foundation models.

In the realm of visual generative modeling, recent works have pioneered the application of Muon to Diffusion Transformers (DiTs) \cite{peebles2023scalable}, such as in pixel MeanFlow \cite{lu2026one} and HunyuanVideo \cite{wu2025hunyuanvideo}. While these initial efforts report an approximate $2\times$ acceleration in convergence speed during early training epochs, empirical observations also reveal a notable plateau effect: although Muon exhibits a sharply decreasing loss initially, the final Fréchet Inception Distance (FID) often merely matches that of AdamW \cite{zhou2025guiding}. In fully trained regimes, the definitive advantage of Muon remains ambiguous; it is unclear whether the optimizer can strictly surpass the generative quality of AdamW or merely reach the same sub-optimal baseline faster.

\begin{figure*}[t!]
\centering
\includegraphics[width=4.8in]{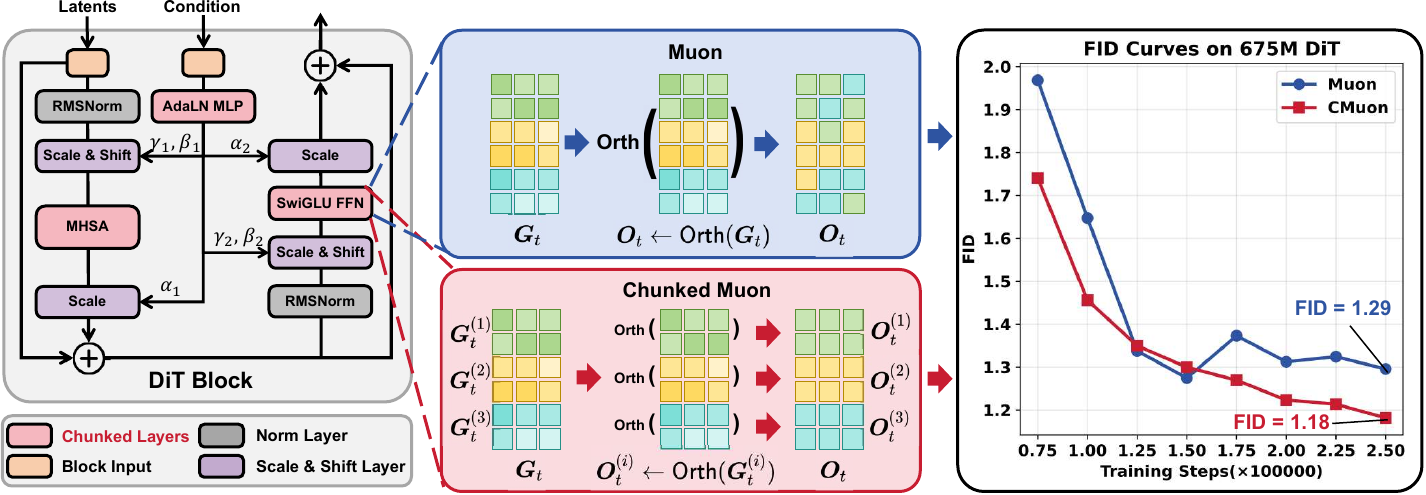}
\caption{\textbf{Overview and empirical validation of CMuon.} \textit{Left:} CMuon applies momentum orthogonalization in a {decoupled} manner by chunking selected weight matrices in Diffusion Transformers (DiTs), replacing a single global orthogonalization with per-chunk orthogonalization. This design mitigates cross-block interference, improving optimization stability and efficiency relative to vanilla Muon. \textit{Mid:} Illustration of our proposed CMuon. \textit{Right:} On ImageNet at $256\times256$, CMuon yields a substantial performance gain over vanilla Muon when training a 675M-parameter DiT model.}
\label{fig:cmuon-main}
\end{figure*}

In this paper, we demonstrate that applying Muon naively to DiTs yields suboptimal results, failing to provide a sustained, definitive speedup over AdamW. We identify the root cause of this bottleneck as the architectural implementation of specific DiT components, namely the AdaLN, QKV, and FFN layers. In standard implementations, functionally distinct weights with independent initializations are concatenated into unified tensors for computational efficiency. Because vanilla Muon applies its Newton-Schulz orthogonalization iteratively to the entire concatenated matrix collectively, it inadvertently induces implicit coupling among these functionally disjoint sub-matrices. This parameter coupling distorts the update directions and ultimately degrades global optimization efficiency. To address this, we propose \textbf{{Chunked Muon (\method)}}, a simple yet highly effective algorithmic modification. By chunking these concatenated matrices back into their original functional sub-components and orthogonalizing their momentums independently, \method fundamentally circumvents the subspace interference issue. This approach dramatically boosts convergence speed up to $2\times$ over AdamW and, crucially, maintains this trajectory advantage throughout the later stages of training.

Furthermore, we conduct comprehensive ablation studies on key optimization hyperparameters, including matrix scaling factors, the selection of layers reserved for AdamW (e.g., embeddings and final projection layers), and the strategy for learning rate scaling. Under the optimal configuration, our 675M-parameter DiT model \cite{peebles2023scalable,sun2025unified}, equipped with the VA-VAE latent space \cite{yao2025reconstruction}, achieves a SOTA FID of 1.18 on ImageNet $256 \times 256$ at 200 epochs (with a batch size of 1024). Notably, we attain this performance solely via the \method optimizer, without relying on auxiliary training acceleration techniques such as Representation Alignment (REPA) \cite{yu2024representation}.
\textbf{In summary, our contributions are threefold:}
\begin{enumerate}[label=(\alph*), nosep, leftmargin=16pt]
    \item We reveal that naively applying Muon to DiTs yields suboptimal convergence, and we trace this bottleneck to the unintended orthogonalization coupling of functionally disjoint parameters concatenated within AdaLN and QKV layers.
    \item To alleviate this limitation, we introduce \method, a simple and effective algorithm that chunks these unified tensors prior to orthogonalization. This eliminates implicit cross-subspace coupling, requires only minor code modifications in the optimizer, and introduces negligible computational overhead.
    \item Extensive experiments demonstrate that \method significantly accelerates the FID reduction rate compared to both AdamW and vanilla Muon. For a 675M DiT model, we achieve a $2\times$ training speedup—an advantage that persists throughout the entire training process. Furthermore, our ablation studies verify the robustness of \method, showing consistent improvements over vanilla Muon across various hyperparameter configurations.
\end{enumerate}

\section{Related Works}

\subsection{Diffusion Transformers for Image Generation}

The integration of transformer architectures into diffusion models~\cite{peebles2023scalable, vaswani2017attention} has redefined the frontier of visual generative modeling, significantly elevating the upper bound of synthesis quality and architectural scalability. This breakthrough has spurred the development of highly efficient variants~\cite{yao2025reconstruction, wang2025ddt, ma2024sit}. To manage the computational demands of high-resolution image generation, modern approaches predominantly adopt the latent diffusion paradigm~\cite{rombach2022high, xie2024sana, kingma2013auto}. By utilizing advanced Variational Autoencoders (VAEs)---such as SD-VAE~\cite{rombach2022high}, VA-VAE~\cite{yao2025reconstruction}, and REPAE~\cite{leng2025repa}---images are compressed into continuous, lower-dimensional latent spaces. Performing diffusion or flow-based modeling within this latent manifold drastically reduces computational overhead while preserving, and often enhancing, reconstruction fidelity.

Formally, the generative process is cast as progressively denoising a sample from a tractable prior distribution along predefined trajectories~\cite{ronneberger2015u}. This dynamics is typically modeled via probability-flow ODEs or flow matching frameworks~\cite{liu2022flow, lipman2022flow, sun2025unified}, with latent trajectories efficiently integrated using DDIM-style discretizations or specialized ODE solvers~\cite{song2020denoising}. Today, this VAE-coupled diffusion framework serves as the de facto standard for open-domain image synthesis~\cite{kingma2013auto, song2020denoising}, forming the backbone of contemporary SOTA open-source models~\cite{wu2025qwen, cai2025z, team2025longcat, batifol2025flux}. Consequently, developing optimization strategies to train these massive Diffusion Transformers rapidly and stably has emerged as a fundamentally critical challenge.

\subsection{Momentum Orthogonalization for Optimization}

The introduction of the Momentum Orthogonalization (Muon) optimizer~\cite{jordan2024muon} has catalyzed a paradigm shift toward optimization strategies based on Linear Minimization Oracles (LMOs). A vibrant line of follow-up research has proposed algorithmic extensions, including Scion~\cite{pethick2025training}, Gluon~\cite{riabinin2025gluon}, and PolarExpress~\cite{amsel2025polar}, which further refine orthogonalization operators and enhance numerical stability. The empirical superiority of this optimizer family has been systematically validated in comprehensive benchmarking efforts; notably, recent studies~\cite{wen2025fantastic, semenov2025benchmarking} demonstrate Muon's advantages across diverse architectures scaling up to one billion parameters. Furthermore, its scalability and efficiency have been rigorously proven in large-scale industrial settings~\cite{team2025kimi, liu2025muon}. Given that Muon has rapidly become an indispensable tool for Large Language Model (LLM) pre-training, extending its success to other foundation models presents a highly promising research trajectory.

\subsection{Muon for DiT Training}

Although initially popularized for accelerating LLM pre-training, Muon has recently emerged as a highly compelling alternative to AdamW for training Diffusion Transformers. The heavy reliance of DiTs on dense projection matrices---specifically within self-attention and multi-layer perceptron blocks---makes them structurally primed for Muon's 2D-matrix orthogonalization operations. Recent optimization benchmarks~\cite{schaipp2025optimization} have empirically validated this synergy, demonstrating that Muon achieves significantly lower final losses and accelerated convergence when learning denoising flow trajectories.

The practical scalability of Muon in visual generative modeling is most prominently showcased by recent large-scale foundation models. For instance, in the 8.3-billion-parameter HunyuanVideo 1.5~\cite{wu2025hunyuanvideo}, Muon was deployed across a multi-stage progressive training regimen spanning text-to-image, text-to-video, and image-to-video tasks. Empirical observations indicate that Muon attains a lower training loss than AdamW in only half the optimization steps, drastically accelerating convergence while preserving high-fidelity motion coherence. Similarly, recent pixel MeanFlow~\cite{lu2026one} highlights that Muon exhibits significantly faster convergence than AdamW when pre-training few-step pixel-space DiTs from scratch. However, despite these early-stage accelerations, applying Muon naively to DiTs often yields suboptimal late-stage convergence—a bottleneck rooted in architectural parameter coupling, which our proposed \method directly addresses.

\section{Methodology}
We study how to stably and efficiently train Diffusion Transformers ~\cite{peebles2023scalable} with Muon-style momentum orthogonalization. While Muon can accelerate convergence and reduce optimizer-state overhead, a practical DiT implementation often {{fuses}} multiple parameter blocks (e.g., QKV projections or AdaLN projections) into a larger 2D matrix for efficiency. This seemingly innocuous concatenation can be problematic: heterogeneous blocks may exhibit substantially different gradient statistics, yet Muon constructs a {shared} preconditioner for the fused matrix, implicitly coupling otherwise unrelated subspaces. We formalize this phenomenon as \textit{subspace interference} in Sec.~\ref{sec:subspace-interference} and show it can distort the descent geometry. To mitigate it, we introduce \textbf{Chunked Muon (\method)} in Algorithm~\ref{alg:muon-step}

\subsection{Preliminaries on Diffusion Models}
\label{sec:prelim-diffusion}
Diffusion-based generative models map a tractable prior distribution to a complex empirical data distribution by simulating a dynamic transport process. While classical diffusion formulations learn to reverse a stochastic noise corruption process~\cite{ho2020denoising,song2020score}, this work adopts the continuous-time Flow Matching (FM) framework~\cite{lipman2022flow}. FM provides a generalized, simulation-free objective for training continuous normalizing flows and has proven highly effective for high-fidelity image generation.
Specifically, let $\mathbf{x} \sim p_{\mathrm{data}}$ denote a target data sample and $\mathbf{z} \sim \mathcal{N}(\bm{0}, \bm{I})$ represent standard Gaussian noise. Flow matching defines a deterministic, probability-flow interpolation path between the data and noise distributions. A standard choice is a simple linear interpolation from data at $t=0$ to noise at $t=1$:
\begin{align}
\mathbf{x}_t \;=\; (1-t)\,\mathbf{x} \;+\; t\,\mathbf{z}, \qquad t\in[0,1] \,.
\label{eq:fm_path}
\end{align}
The time derivative of this path yields a constant-velocity target vector field, $\mathbf{z} - \mathbf{x}$. Given an optional conditioning signal $\mathbf{c}$ (e.g., text embeddings), we train a neural network $\bm{F}_{\bm{\theta}}(\mathbf{x}_t, t, \mathbf{c})$ to regress this target vector field via a straightforward squared-error objective:
\begin{align}
\mathcal{L}(\bm{\theta})
\;=\;
\mathbb{E}_{\mathbf{x},\mathbf{z},\mathbf{c},t}
\Big[
\big\|\bm{F}_{\bm{\theta}}(\mathbf{x}_t,t,\mathbf{c})-(\mathbf{z}-\mathbf{x})\big\|_2^2
\Big] \,.
\label{eq:fm_loss}
\end{align}
During inference, generation is framed as an initial value problem (IVP). Starting from a noise sample $\mathbf{x}_1 \sim \mathcal{N}(\bm{0}, \bm{I})$, we generate a clean data sample $\mathbf{x}_0$ by integrating the learned vector field backward in time (from $t=1$ to $t=0$) using the corresponding probability-flow ordinary differential equation (ODE):
\begin{align}
\frac{\mathrm{d}\mathbf{x}_t}{\mathrm{d}t} \;=\; \bm{F}_{\bm{\theta}}(\mathbf{x}_t,t,\mathbf{c}) \,.
\label{eq:pf_ode}
\end{align}

\subsection{The Muon Optimizer}
\label{sec:muon-optimizer}
Under the diffusion training setup above, we optimize model parameters $\bm{\theta}$ by minimizing the flow-matching objective in Eq.~\eqref{eq:fm_loss}:
\begin{align}
\bm{\theta}^\star
\;=\;
\arg\min_{\bm{\theta}}\;
\mathcal{L}(\bm{\theta}),
\qquad
\mathcal{L}(\bm{\theta})
=
\mathbb{E}_{\mathbf{x},\mathbf{z},\mathbf{c},t}
\Big[
\big\| \bm{F}_{\bm{\theta}}(\mathbf{x}_t,t,\mathbf{c})-(\mathbf{z}-\mathbf{x}) \big\|_2^2
\Big].
\label{eq:diff_opt_problem}
\end{align}
Standard optimizers  view the gradient $\nabla_{\bm{\theta}}\mathcal{L}$ as a collection of vectors. In contrast, Muon~\cite{jordan2024muon} operates {matrix-wise}: for each 2D weight tensor (e.g., a linear projection) $\bm{W}\in\mathbb{R}^{m\times n}$, Muon constructs an update direction by {orthogonalizing} the momentum matrix. This prevents rank collapse of the update by effectively replacing the singular-value spectrum with an identity while removing anisotropic scaling in principle subspaces.

Let $\bm{W}_t\in\mathbb{R}^{m\times n}$ be a layer weight at iteration $t$, and let $\bm{G}_t=\nabla_{\bm{W}}\mathcal{L}(\bm{\theta}_t)\in\mathbb{R}^{m\times n}$ denote its stochastic gradient computed from minibatches of $(\mathbf{x},\mathbf{z},\mathbf{c},t)$ sampled as in Sec.~\ref{sec:prelim-diffusion}. Muon maintains a momentum buffer $\bm{M}_t\in\mathbb{R}^{m\times n}$ and performs
\begin{align}
\bm{M}_t
&=
\mu\,\bm{M}_{t-1} + \bm{G}_t,
\label{eq:muon_momentum}\\
\bm{O}_t
&=
\mathrm{Orth}(\mu\bm{M}_t + \bm{G}_t),
\label{eq:muon_ortho}\\
\bm{W}_t
&=
\bm{W}_{t-1} ( 1-\eta\lambda) - \eta  \cdot 0.2 \sqrt{\max(m, n)} \bm{O}_t,
\label{eq:muon_param_update}
\end{align}
where $\eta$, $\mu$ and $\lambda$ are the learning rate, momentum coefficient and weight decay rate, respectively, and $0.2\sqrt{\max(m,n)}$ is a dimensional scaling factor that improves scalability~\cite{liu2025muon}. Note that we use Nesterov-type momentum~\cite{nesterov1983method}. The operator $\mathrm{Orth}(\cdot)$ maps an input matrix to its closest semi-orthogonal matrix in the sense of the polar factor. Concretely, if the  SVD of $\bm{M}$ is $\bm{M}=\bm{U}\bm{\Sigma}\bm{V}^\top$, then
\begin{align}
\mathrm{Orth}(\bm{M}) \;:=\; \bm{U}\bm{V}^\top.
\label{eq:ortho_def}
\end{align}
In practice, Muon avoids explicit SVD by approximating $\bm{U}\bm{V}^\top$ via Newton--Schulz iterations for the inverse square root. Several variants further improve numerical accuracy and speed~\cite{amsel2025polar}.

\subsection{Subspace Interference in Muon}\label{sec:subspace-interference}

In DiT, many independent parameters are initialized within the same module for efficiency (e.g., {scale} projection and {shift} projection in AdaLN). However, the projection structures and gradient statistics of these parameter blocks can differ substantially. When Muon is applied to a concatenation of such heterogeneous blocks, the orthogonalization step can induce unexpected \textbf{coupling} across otherwise unrelated subspaces.

Consider a toy example with $N$ gradient submatrices $\bm{G}_1,\dots,\bm{G}_N \in \mathbb{R}^{d\times d}$ stacked into a tall matrix
\begin{align}
\bm{G} \in \mathbb{R}^{Nd \times d}, \qquad 
\bm{G} = [\bm{G}_1,\bm{G}_2,\cdots,\bm{G}_N]^\top .
\label{eq:stacked_grad}
\end{align}
Muon approximates an orthogonalized update through Newton--Schulz iterations:
\begin{align}
\bm{U} \approx \bm{G}(\bm{G}^\top \bm{G})^{-1/2}.
\label{eq:muon_update}
\end{align}
\noindent\textbf{Vanilla Muon (coupled).}
If Muon is applied to the stacked matrix $\bm{G}$ in Eq.~\eqref{eq:stacked_grad}, the update follows Eq.~\eqref{eq:muon_update} with
\begin{align}
\bm{U}_i' = \bm{G}_i
\left(\sum_{j=1}^{N}\bm{G}_j^\top \bm{G}_j\right)^{-1/2}.
\label{eq:coupled_update}
\end{align}
\noindent\textbf{Chunked Muon (decoupled).}
If Muon is instead applied independently to each block, the $i$-th update direction becomes
\begin{align}
\bm{U}_i = \bm{G}_i(\bm{G}_i^\top \bm{G}_i)^{-1/2}.
\label{eq:chunked_update}
\end{align}
Here the preconditioner $(\bm{G}_i^\top \bm{G}_i)^{-1/2}$ depends only on the statistics of $\bm{G}_i$.
Comparing Eq.~\eqref{eq:chunked_update} and Eq.~\eqref{eq:coupled_update}, we observe that stacked Muon replaces the block-specific preconditioner with a shared one $(\bm{G}^\top \bm{G})^{-1/2}$. This shared preconditioner mixes gradient covariance structures across blocks that may correspond to unrelated roles. When the dominant principal directions of $\bm{G}_1$ and $\bm{G}_2$ are misaligned, the summed covariance distorts each block's preferred descent geometry, producing severe subspace interference. Further theoretical discussions and analyses of \method are provided in Appendix~\ref{sec:further-theory}.





\subsection{Algorithm Design}\label{sec:alg_design}
Our algorithm is intentionally simple yet efficient. For parameter blocks that exhibit cross-subspace coupling under fused Muon, we partition each fused matrix into multiple {chunks} and perform orthogonalization {independently} on each chunk. Following prior DiT implementations \cite{yao2025reconstruction, sun2025unified}, the fused matrices that most benefit from chunking fall into three categories: (i) AdaLN (Adaptive LayerNorm) modulation, (ii) attention QKV projections, and (iii) FFN gate+up projections.

Table~\ref{tab:chunk-config} summarizes our chunking scheme for DiT-XL with hidden dimension $d=1152$. In all cases, we split along the longer dimension so that each chunk corresponds to a semantically independent sub-matrix before applying Newton--Schulz orthogonalization.

\begin{table}[t]
\centering
\caption{Chunk configuration for fused linear layers in DiT-XL ($d = 1152$). Each fused weight matrix is split along the longer dimension into semantically independent sub-matrices before Newton--Schulz orthogonalization.}
\label{tab:chunk-config}
\resizebox{\linewidth}{!}{%
\begin{tabular}{lcccc}
\toprule
\textbf{Layer Type} & \textbf{Original Shape} & \textbf{Chunk Dim.} & \textbf{\#Chunks} & \textbf{Per-Chunk Shape} \\
\midrule
QKV Projection   & $[3456,\;1152]$ & 0 & 3 & $[1152,\;1152]$ \\
MLP Gate+Up      & $[6144,\;1152]$ & 0 & 2 & $[3072,\;1152]$ \\
AdaLN Modulation & $[6912,\;1152]$ & 0 & 6 & $[1152,\;1152]$ \\
\bottomrule
\end{tabular}%
}
\end{table}

\subsubsection{CMuon update.}

Algorithm~\ref{alg:muon-step} formalizes one optimization step of our \method. For each parameter tensor, we (i) apply momentum to the raw gradient, (ii) split into chunks and orthogonalize each chunk via Newton--Schulz iterations, and (iii) apply the scaled update. Here we omit the detailed reshape operation for concise expression.

\begin{algorithm}[b!]
\caption{Chunked Muon Optimizer Step}
\label{alg:muon-step}
\DontPrintSemicolon
\SetKwInOut{Input}{Input}

\Input{Tuple set $\mathcal{P}_t=\{(\bm{W}_t,\bm{G}_t,d_{{\mathrm{chunk}}},N_{{\mathrm{chunk}}},\alpha)\}$ over all Muon parameters, where $(d_{{\mathrm{chunk}}},N_{{\mathrm{chunk}}},\alpha)$ is initialized by Algorithm~\ref{alg:muon-init}; learning rate $\eta$, momentum $\mu$, weight decay $\lambda$, and Newton--Schulz iterations $K$.}

Define $\mathrm{NewtonSchulz}(\cdot,\cdot)$ in Algorithm~\ref{alg:newton-schulz}.\;

\For{$t=1,2,\ldots$}{
  \ForEach{$(\bm{W}_t,\bm{G}_t,d_{{\mathrm{chunk}}},N_{{\mathrm{chunk}}},\alpha)\in\mathcal{P}_t$}{
    $\bm{M}_t \leftarrow \mu\,\bm{M}_{t-1} + \bm{G}_t$.\;
    $\bm{G}_t \leftarrow \bm{G}_t + \mu\,\bm{M}_t$.\;

    \eIf{$N_{{\mathrm{chunk}}} > 1$}{
      Split $\bm{G}_t$ along dimension $d_{{\mathrm{chunk}}}$ into $\{\bm{G}_t^{(i)}\}_{i=1}^{N_{{\mathrm{chunk}}}}$.\;
      \For{$i=1,\ldots,N_{{\mathrm{chunk}}}$}{
        $\bm{O}_t^{(i)} \leftarrow \mathrm{NewtonSchulz}(\bm{G}_t^{(i)}, K)$.\;
      }
      $\bm{O}_t \leftarrow \mathrm{Concat}\!\big(\bm{O}_t^{(1)},\ldots,\bm{O}_t^{(N_{{\mathrm{chunk}}})}\big)$ along dimension $d_{{\mathrm{chunk}}}$.\;
    }{
      $\bm{O}_t \leftarrow \mathrm{NewtonSchulz}(\bm{G}_t, K)$.\;
    }
    
    
    


    $\bm{W}_t \leftarrow (1-\eta\lambda)\,\bm{W}_t \;-\; \eta\,\alpha\,\bm{O}_t$.\;
  }
}
\end{algorithm}
\begin{algorithm}[b!]
\caption{Chunked Muon Initialization}
\label{alg:muon-init}
\DontPrintSemicolon
\SetKwInOut{Input}{Input}

\Input{Muon parameter set $\{\bm{W}^{(j)}\}$ with associated names; chunking configuration in Table~\ref{tab:chunk-config}; rescaling flag $r\in\{\mathrm{False},\mathrm{True}\}$.}

Initialize momentum buffers $\bm{M}^{(j)}_{0} \leftarrow \bm{0}$ for each parameter $\bm{W}^{(j)}$.\;

\ForEach{Muon parameter $\bm{W}^{(j)}$ with 2-dimensional shape}{
    Determine $(d_{\mathrm{chunk}}, N_{\mathrm{chunk}})$ according to Table~\ref{tab:chunk-config} and obtain shape $(d_{\mathrm{out}}, d_{\mathrm{in}})$ after chunking.\;
    $\alpha^{(j)} \leftarrow 0.2\sqrt{\max(d_{\mathrm{out}}, d_{\mathrm{in}})}$.\;
    \If{$r = \mathrm{True}$}{
        $\alpha^{(j)} \leftarrow \alpha^{(j)} \cdot \sqrt{N_{\mathrm{chunk}}}$.\;
    }
    Store $(d_{\mathrm{chunk}},\, N_{\mathrm{chunk}},\, \alpha^{(j)})$ in the optimizer state of $\bm{W}^{(j)}$.\;
}
\end{algorithm}

\subsubsection{Learning-rate adjustment under chunking.}\label{sec:alg_design}

To balance the update scale between blocks using AdamW and Muon, we adopt the Moonlight variant of Muon scaling, which allows us to directly reuse hyperparameters from AdamW. When chunking is enabled, we recompute the scaling factor based on the per-chunk dimensions so that the {overall} Frobenius norm of the stacked update remains unchanged. In this way, chunking only redistributes the norm across chunks.  

Specifically, let a fused 2D gradient be $\bm{G}\in\mathbb{R}^{(N d_{\mathrm{out}})\times d_{\mathrm{in}}}$ with $d_{\mathrm{out}} \ge d_{\mathrm{in}}$, where $N$ denotes the number of chunks and each chunk $\bm{G}_i\in\mathbb{R}^{d_{\mathrm{out}}\times d_{\mathrm{in}}}$. Under Moonlight scaling,
\begin{align}
\alpha &= 0.2\sqrt{\max(Nd_{\mathrm{out}},\, d_{\mathrm{in}})}
      = 0.2\sqrt{Nd_{\mathrm{out}}}, 
\label{eq:alpha_full}
\\
\alpha_c &= 0.2\sqrt{\max(d_{\mathrm{out}},\, d_{\mathrm{in}})}
        = 0.2\sqrt{d_{\mathrm{out}}}.
\label{eq:alpha_chunk}
\end{align}
Let $\mathrm{Orth}(\cdot)$ denote the polar factor in \eqref{eq:ortho_def} (approximated by Newton--Schulz in Algorithm~\ref{alg:newton-schulz}). 
For $m\ge n$, $\mathrm{Orth}(\bm{G})$ has orthonormal columns, hence $\|\mathrm{Orth}(\bm{G})\|_F=\sqrt{n}=\sqrt{d_{\mathrm{in}}}$.
The Frobenius norm of the fused update becomes
\begin{align}
\left\|\alpha\,\mathrm{Orth}(\bm{G})\right\|_F
= \alpha\sqrt{d_{\mathrm{in}}}
= 0.2\sqrt{Nd_{\mathrm{out}}d_{\mathrm{in}}}.
\label{eq:norm_full}
\end{align}
where $\mathrm{Orth}(\cdot)$ is defined in \eqref{eq:ortho_def}.

For the chunked counterpart obtained by stacking the chunk updates,
\begin{align}
\Bigl\|\alpha_c\,
[\mathrm{Orth}(\bm{G}_1);
 \mathrm{Orth}(\bm{G}_2);
 \cdots;
 \mathrm{Orth}(\bm{G}_N)]
\Bigr\|_F
&= \alpha_c\sqrt{Nd_{\mathrm{in}}}\\
&= 0.2\sqrt{Nd_{\mathrm{out}}d_{\mathrm{in}}}.
\label{eq:norm_chunk}
\end{align}

Comparing Eq.~\eqref{eq:norm_full} and Eq.~\eqref{eq:norm_chunk}, we observe that the Frobenius norm of the full $(Nd_{\mathrm{out}})\times d_{\mathrm{in}}$ update is preserved. Therefore, under the Moonlight formulation, chunking maintains the global update magnitude while only redistributing the norm across chunks.


Furthermore, we find that scaling the learning-rate multiplier of the chunked components by $\sqrt{N_{\mathrm{chunk}}}$ can accelerate early-stage convergence without compromising the final performance. Therefore, we provide an optional rescaling switch $r$ in Algorithm~\ref{alg:muon-init}. Detailed ablation results are discussed in Table~\ref{tab:abl_rescale}.

\section{Experiments}

\subsection{Experimental Setup}

\subsubsection{Models and datasets.}\label{sec:models_datasets}
We conduct experiments on ImageNet-1K~\cite{deng2009imagenet} at a resolution of $256\times256$, following standard preprocessing protocols used in prior diffusion-modeling work~\cite{dhariwal2021diffusion}. All models adopt DiTs ~\cite{peebles2023scalable} as the backbone.\footnote{Following \cite{sun2025unified}, we apply minor architectural modifications for neural network for improved stability and performance.} We perform latent-space generative modeling using VA-VAE~\cite{yao2025reconstruction} as the default autoencoder, and additionally employ SD-VAE~\cite{rombach2022high} in ablation studies to assess the sensitivity of our findings to the choice of VAE. For sampling, we use model guidance to reduce the inference overhead induced by classifier-free guidance~\cite{ho2022classifier,tang2025diffusion}. Unless otherwise specified, all experiments are evaluated with 30 NFEs. 

\subsubsection{Implementation details.}\label{sec:implementation}

Following standard practices, we evaluate EMA models (decay 0.9999) using FID-50K~\cite{heusel2017gans}. Training employs a constant learning rate, a global batch size of 1024, \texttt{bf16} precision, and gradient clipping (max norm 1.0). AdamW parameters are set to $(\beta_1,\beta_2)=(0.9,0.95)$ with 0 weight decay. Crucially, Muon and \method optimize only 2D weights in attention/FFN/AdaLN projections, while AdamW handles 1D parameters, embeddings, and the final layer~\cite{jordan2024muon,wu2025hunyuanvideo}. See Appendix~\ref{sec:exp-details} for further details.

\begin{figure}[t!]
    \centering
    \subfloat[\revision{\textbf{DiT-B.} FID@50k versus training progress for AdamW, Muon, and \method. Muon/\method accelerate early-stage improvement, while \method sustains gains into late training.}]{%
        \includegraphics[width=0.48\textwidth]{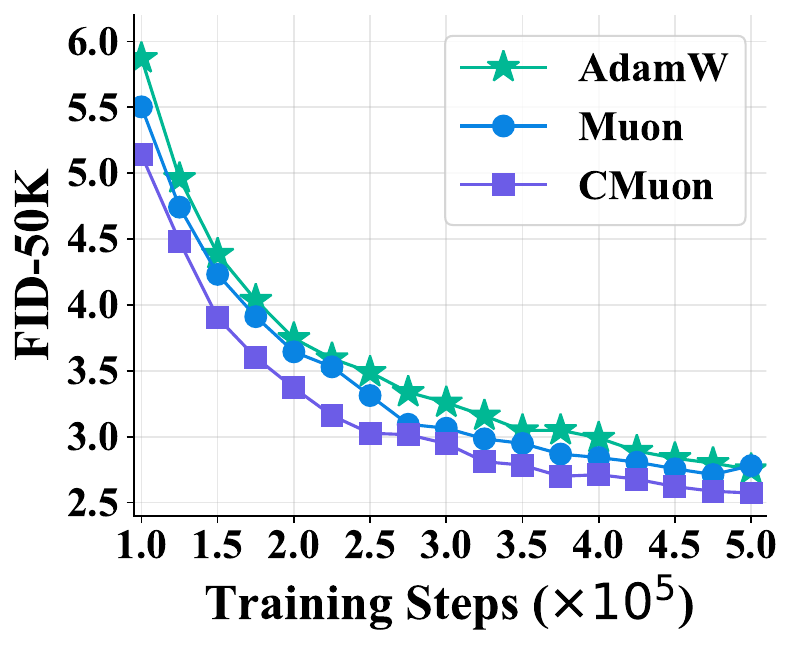}%
        \label{fig:fid-main-b}%
    }
    \hfil
    \subfloat[\revision{\textbf{DiT-XL .} FID@50k versus training progress for AdamW, Muon, and \method. \method maintains continued FID reduction and reaches the same FID up to $2\times$ faster than AdamW.}]{%
        \includegraphics[width=0.48\textwidth]{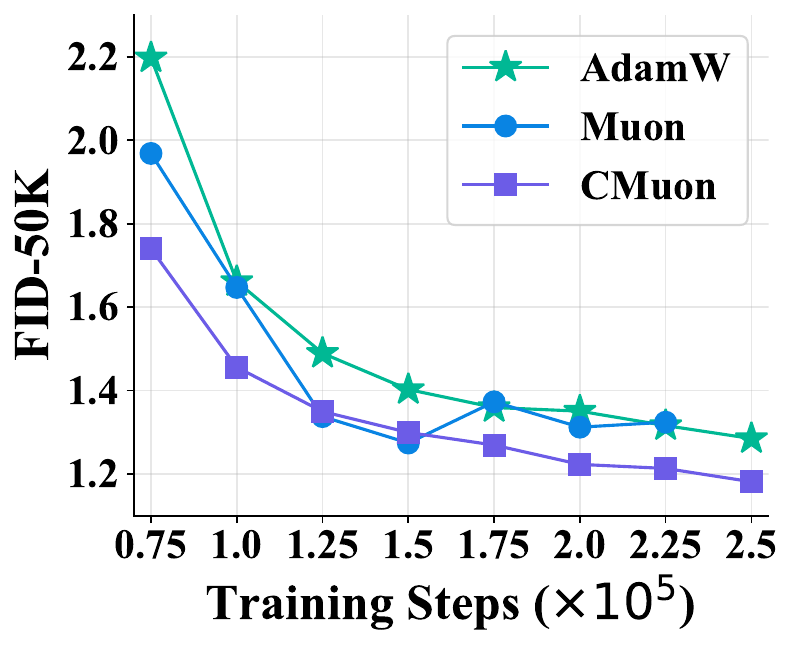}%
        \label{fig:fid-main-xl}%
    }
    \caption{\textbf{DiT training comparison on ImageNet-1K $256\times256$.} FID@50k (lower is better) as a function of training progress for AdamW, Muon, and \method at two model scales. Muon and \method improve markedly faster in early training, while \method further preserves late-stage convergence and achieves better final quality.}
    \label{fig:fid-main}
\end{figure}

\begin{table*}[t!]
    \centering
    \caption{\small{\textbf{Comparison of optimizers on image generation task on class-conditional ImageNet-1K.} Across settings, vanilla Muon reduces FID faster in the early stage, but its improvement diminishes later and becomes comparable to AdamW by 200 epochs. In contrast, CMuon maintains a faster late-stage convergence than AdamW while achieving higher final accuracy, consistently converging to a lower FID.}}
    \vspace{-0.5em}
    \label{tab:fid-main}
    \begin{tabular}{lccccccc}
        \toprule
        \multirow{2}{*}{\textbf{Optimizer}} &
        \multirow{2}{*}{\textbf{Model}} &
        \multirow{2}{*}{\textbf{\#Params}} &
        \multirow{2}{*}{\textbf{VAE}} &
        \multirow{2}{*}{\textbf{NFE} ($\downarrow$)} &
        \multicolumn{3}{c}{\textbf{FID} ($\downarrow$)} \\
        \cmidrule(lr){6-8}
        & & & & & \textbf{ep80} & \textbf{ep200} & \textbf{ep400} \\
        \midrule
        AdamW    & DiT-B  & 130M & VA-VAE & $30$ & 5.87  & 3.49 & 2.78 \\
        Muon     & DiT-B  & 130M & VA-VAE & $30$ & 5.50  & 3.31 & 2.78 \\
        \method  & DiT-B  & 130M & VA-VAE & $30$ & 5.14  & 3.03 & 2.57 \\
        \midrule
        AdamW    & DiT-B  & 130M & SD-VAE & $30$ & 5.87  & 3.60 & -- \\
        Muon     & DiT-B  & 130M & SD-VAE & $30$ & 5.39  & 3.30 & -- \\
        \method  & DiT-B  & 130M & SD-VAE & $30$ & 4.94 & 3.06 & -- \\
        \midrule
        AdamW    & DiT-XL & 675M & VA-VAE & $30$ & 1.66 & 1.30 & 1.21 \\
        Muon     & DiT-XL & 675M & VA-VAE & $30$ & 1.65 & 1.29 & --   \\
        \method  & DiT-XL & 675M & VA-VAE & $30$ & \textbf{1.46} & \textbf{1.18} & --   \\
        \bottomrule
    \end{tabular}
    \vspace{-1em}
\end{table*}

\subsection{Performance of \method in DiT Training}

We evaluate \method against AdamW and vanilla Muon on DiT-Base  and DiT-XL . As Figure~\ref{fig:fid-main} shows, while both Muon variants accelerate early-stage training, vanilla Muon plateaus in later stages. In contrast, \method sustains its convergence rate throughout training. Ultimately, \method achieves an FID of 1.18 in just 200 epochs, outperforming AdamW’s 400-epoch FID of 1.21 (Table~\ref{tab:fid-main}).
Overall, \method delivers over a 2x training speedup compared to AdamW.

\begin{figure}[t!]
    \centering
    \subfloat[\revision{\textbf{Muon} after training 200 epochs, FID@50k = 1.29.}]{%
        \includegraphics[width=0.48\textwidth]{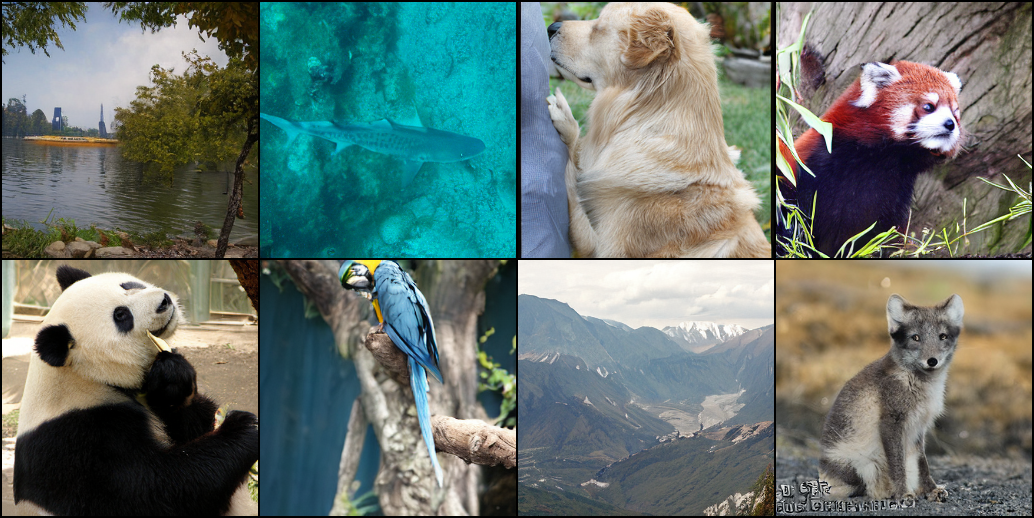}%
    }
    \hfil 
    \subfloat[\revision{\textbf{CMuon} after training 200 epochs, FID@50k = 1.18.}]{%
        \includegraphics[width=0.48\textwidth]{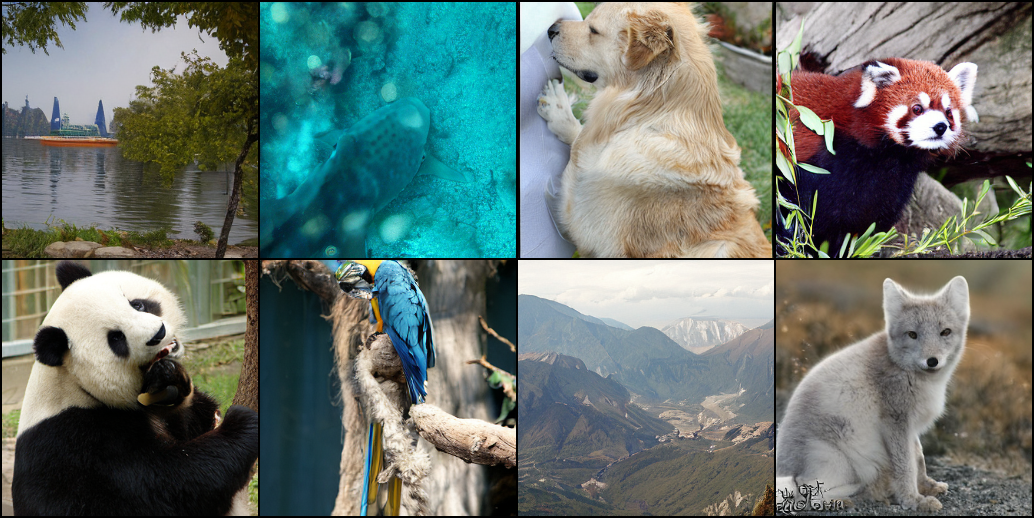}%
    }
    \caption{\textbf{Qualitative comparison on ImageNet-1K $256\times256$.} Random class-conditional samples generated by DiT-XL  trained with Muon (left) and \method (right). All images are sampled with the same sampling configuration and $\mathrm{NFE}=30$.}
    \label{fig:fid-sample}
\end{figure}

Figure~\ref{fig:fid-sample} provides a qualitative comparison. Given the same DiT-XL setup, 200-epoch budget, and 30 NFE, \method yields visually cleaner and more coherent images than vanilla Muon. Specifically, \method better captures fine details and mitigates common failures like texture artifacts and structural distortions. Aligning with our FID results, this confirms that chunked optimization improves not only early-stage speed but also final perceptual quality.

\section{Further Ablation Studies}



\subsection{Ablation on Blocks for Chunking}

To more precisely validate the importance of applying chunking to different architectural blocks, we conduct an ablation study on the 130M DiT-B model by selectively enabling chunking for different projection layers. In this experiment, {+FFN} denotes applying chunking to the gate and up projections in the feed-forward network (FFN), {+QKV} denotes chunking the query, key, and value projections in the multi-head attention module, and {+AdaLN} refers to chunking the projection layer inside Adaptive LayerNorm. Under this setting, {None} corresponds to the original Muon optimizer without chunking, while the other configurations represent variants of our \method with different block-level chunking strategies. 

We report the FID after 80 and 200 training epochs to evaluate both early-stage convergence and final generative performance. As shown in Table~\ref{tab:abl_chunked_block}, applying chunking to any single block individually provides only marginal improvements in the later stage of training, reducing the final FID by at most 0.1 compared with the baseline. Nevertheless, a modest acceleration in convergence can be observed for several configurations. In contrast, when chunking is jointly applied to FFN, QKV, and AdaLN projections, the improvement becomes substantially more pronounced. Specifically, the final FID decreases from 3.31 to 3.02 at 200 epochs, corresponding to an improvement of approximately 0.3, while the early-stage performance at 80 epochs also shows a clear gain. These results suggest that chunking different architectural blocks produces complementary optimization benefits, and jointly applying chunking to FFN, QKV, and AdaLN yields the most effective and balanced configuration for stabilizing and accelerating Muon-based training.

\begin{table}[t!]
    \centering
    \caption{\small{\textbf{Ablation Study on Chunked Blocks: Impact of QKV, FFN, and AdaLN Projections on 130M DiT-B Models.}}   The table presents the performance of the 130M DiT-B model on the FID metric at 80 and 200 epochs under different configurations of chunked blocks. The results show that the inclusion of FFN, QKV, and AdaLN projections leads to a notable improvement in FID at both epochs compared to other individual components, highlighting the importance of combining these elements for enhanced model performance.}
        \label{tab:abl_chunked_block}
    \begin{tabular}{lcc}
        \toprule
        \textbf{Chunked Blocks}          & \textbf{FID@80ep}   & \textbf{FID@200ep} \\
        \midrule
        None                             &   5.50                & 3.31                      \\
        FFN                            &   5.43                & 3.28                      \\
        QKV                             &   5.32                & 3.35                      \\
        AdaLN                           &   6.00                & 3.23                      \\
        FFN, QKV, AdaLN                 &   \textbf{5.14}                & \textbf{3.02}                     \\
        \bottomrule
    \end{tabular}
    \vspace{-1em}
\end{table}







\subsection{Effect of Learning Rate Scaling Strategies}


When transitioning from AdamW-style optimization to Muon, several block-wise learning rate scaling strategies have been proposed to stabilize training across different parameter shapes. To evaluate the robustness of \method under different scaling rules, we conduct an ablation study using three representative strategies: {Vanilla} (no learning rate scaling), the {Keller--Jordan} strategy, and the {MoonLight} strategy. The detailed scaling configurations are summarized in Table~\ref{tab:abl_lr_scaler}. In particular, $d_{\mathrm{out}}$ and $d_{\mathrm{in}}$ denote the output and input dimensions of a parameter matrix after reshaping it into a two-dimensional form. Following our standard experimental protocol, all models are trained on the 130M DiT-B architecture for 200 epochs, and the final FID scores are reported in Table~\ref{tab:abl_lr_scaler}.

The results demonstrate that \method consistently achieves lower FID scores than the original Muon optimizer across all three scaling strategies, indicating that the proposed chunking mechanism improves optimization stability regardless of the specific learning rate scaling rule. It is also worth noting that the {Vanilla} and {Keller--Jordan} configurations produce noticeably worse FID compared with the {MoonLight} strategy. This behavior mainly arises because we directly adopt the learning rate originally tuned for AdamW without further retuning, which leads to suboptimal scaling for the Vanilla and Keller--Jordan rules. In contrast, the MoonLight scaling naturally aligns better with this learning rate choice and therefore yields significantly better performance. For this reason, we adopt the MoonLight scaling strategy as the default configuration in the remainder of this paper.

\begin{table}[t]

    \centering
    \caption{\small{\textbf{Ablation study on Muon learning rate scaling strategies for DiT-B, with all models trained for 200 epochs.}This table compares the performance of different learning rate scaling strategies, including Vanilla, KellerJordan, and MoonLight, on both Muon and \method  models. The scaling strategies are evaluated based on their respective FID scores, with MoonLight showing the best performance in both cases.}}
        \label{tab:abl_lr_scaler}
    \begin{tabular}{lccc}
        \toprule
         \textbf{Scale Function} & \textbf{Vanilla} & \textbf{KellerJordan} & \textbf{MoonLight} \\
         \textbf{Scaling factor}   & $\times 1$       & $\times \max\!\left(\sqrt{\frac{d_{\mathrm{out}}}{d_{\mathrm{in}}}},\, 1\right)$ & $\times 0.2\sqrt{\max(d_{\mathrm{out}},d_{\mathrm{in}})}$ \\
        \midrule
         Muon&    8.73              &    7.40 &                    \textbf{3.31}\\
         \method&   6.94               & 6.00           &\textbf{3.03}\\
         \bottomrule
    \end{tabular}
    \vspace{-1em}
\end{table}

\subsection{Ablation Study on Learning Rate Rescaling}

\begin{table}[b!]
    \centering
    \caption{\small{\textbf{Ablation study on learning-rate rescaling for chunked layers.} This table explores the effect of varying learning-rate rescaling on chunked layers in the Muon and \method optimizers. We observe the influence of different rescaling strategies on the FID scores at both 40 and 80 epochs, showing significant improvements when rescaling is applied based on chunk size.}}
    \label{tab:abl_rescale}
    \begin{tabular}{lcccc}
        \toprule
        \textbf{Optimizer} & \textbf{Base LR Scale} & \textbf{Chunk Rescale} & \textbf{FID@40ep} & \textbf{FID@80ep} \\
        \midrule
        Muon  & $0.2\sqrt{\max(d_{\mathrm{out}},d_{\mathrm{in}})}$ & $\times 1$                      & 5.67 & 1.65 \\
        Muon  & $0.2\sqrt{\max(d_{\mathrm{out}},d_{\mathrm{in}})}$ & $\times \sqrt{N_{\mathrm{{chunk}}}}$ & 4.26 & 1.55 \\
        \method & $0.2\sqrt{\max(d_{\mathrm{out}},d_{\mathrm{in}})}$ & $\times 1$                      & 5.32 & 1.50 \\
        \method & $0.2\sqrt{\max(d_{\mathrm{out}},d_{\mathrm{in}})}$ & $\times \sqrt{N_{\mathrm{chunk}}}$ & \textbf{3.78} & \textbf{1.46} \\
        \bottomrule
    \end{tabular}
    \vspace{-1em}
\end{table}

As mentioned at the end of Section~\ref{sec:alg_design}, we observe that rescaling the learning rate of the chunked blocks by a factor of $\sqrt{N_{\mathrm{chunk}}}$ can substantially improve early-stage convergence without slowing down optimization in the later stages. To isolate the effect of this rescaling trick, we apply it to both Muon and \method and compare their FID scores after 40 and 80 training epochs. For the {Muon + Rescale} setting, we rescale only the same set of layers that are chunked in \method, while keeping these layers {unchunked} and optimized with standard Muon. Such operation ensures that the comparison targets the learning-rate effect rather than the chunking operation itself. All configurations follow the same Scaling Factor MoonLight and the results are summarized in Table~\ref{tab:abl_rescale}.

As shown in Table~\ref{tab:abl_rescale}, learning-rate rescaling and chunking exhibit clear complementarity: each brings consistent gains, and their combination yields the best overall performance. In particular, the two {Rescale} configurations achieve the lowest FID at 40 epochs, indicating that $\sqrt{N_{chunk}}$ rescaling primarily accelerates early diffusion training. By 80 epochs, the configurations using \method deliver the best FID, suggesting that chunking plays the dominant role in improving the attainable final performance. Taken together, these results imply a division of labor between the two techniques: rescaling mainly improves early-stage optimization speed, whereas chunking largely determines the ultimate FID improvements and remains the primary contributor to late-stage gains.

\subsection{Ablation Study on Learning Rate}

\begin{table}[t!]
    \centering
    \caption{\small{\textbf{Analysis on best learning rate and training epochs when training DiT.} We report FID@50k for AdamW and \method across learning rates $\{1,2,3\}\times10^{-4}$ and training lengths of 80/140/200 epochs.}}
    \label{tab:abl_lr_epoch_b}
    \begin{tabular}{lcccccc}
        \toprule
        \multirow{2}{*}{\textbf{LR}} &
        \multicolumn{3}{c}{\textbf{AdamW}} &
        \multicolumn{3}{c}{\textbf{\method}} \\
        \cmidrule(lr){2-4}\cmidrule(lr){5-7}
        & \textbf{FID@80ep} & \textbf{FID@140ep} & \textbf{FID@200ep}
        & \textbf{FID@80ep} & \textbf{FID@140ep} & \textbf{FID@200ep} \\
        \midrule
        $1\mathrm{e}{-4}$ & 2.13 & --   & --   & 1.79 & --   & --   \\
        $2\mathrm{e}{-4}$ & 1.66 & 1.36 & 1.29 & 1.46 & 1.27 & 1.18 \\
        $3\mathrm{e}{-4}$ & 1.63 & 1.32 & 1.26 & 1.54 & 1.25 & 1.20 \\
        \bottomrule
    \end{tabular}
    \vspace{-1em}
\end{table}

To ensure that our comparisons are conducted under near-optimal hyperparameters, we ablate the learning rate when training DiT-XL with 675M parameters. Motivated by the commonly adopted setting in UCGM~\cite{sun2025unified}, we evaluate learning rates in the neighborhood of $2\times10^{-4}$, specifically $\{1,2,3\}\times10^{-4}$, and report the resulting FID@50k at multiple training horizons (80/140/200 epochs) for both AdamW and \method. Table~\ref{tab:abl_lr_epoch_b} summarizes the sensitivity of final image quality to learning rate and training length, enabling a fair assessment of optimizer behavior without confounding from poorly tuned learning rates.

Table~\ref{tab:abl_lr_epoch_b} shows that $2\times10^{-4}$ is consistently near-optimal across both optimizers, achieving strong performance at all evaluated epochs. While AdamW can reach a competitive FID of $1.26$ with a larger learning rate ($3\times10^{-4}$) after 200 epochs, \method attains comparable or better quality substantially earlier: for instance, \method achieves FID $1.27$ at 140 epochs with $2\times10^{-4}$, closely matching AdamW's 200-epoch result, thereby maintaining the convergence speedup observed in our main experiments. Overall, this ablation confirms that our reported gains are not an artifact of an unfavorable learning-rate choice for AdamW, and that \method remains robust and effective under well-searched learning-rate settings.

\section{Conclusion}

We introduce \method to resolve the late-stage convergence plateaus of the Muon optimizer in Diffusion Transformers. Because standard DiTs fuse distinct projections (e.g., QKV, AdaLN), joint orthogonalization causes cross-subspace coupling. \method simply chunks these matrices back into independent sub-components before orthogonalization. On ImageNet-1K $256\times256$, this zero-overhead modification achieves a SOTA FID of 1.18 on DiT-XL  at 200 epochs—surpassing AdamW, which requires 400 epochs to reach 1.21. \method establishes a highly efficient and stable optimization standard for large-scale DiT training.

\bibliographystyle{splncs04}
\bibliography{main.bib}

\newpage
\appendix
\section{Experimental Settings}
\label{sec:exp-details}

\subsection{Detailed Settings of Training and Sampling}

\begin{table}[b!]
\centering
\caption{Model configurations for DiT-B and DiT-XL.}
\label{tab:tit_model_config}
\setlength{\tabcolsep}{3pt}
\begin{tabular}{lccccccc}

\toprule
\textbf{Model} & \textbf{Params} & \textbf{Depth} & \textbf{\renewcommand{\arraystretch}{1.0}\begin{tabular}[c]{@{}c@{}}Hidden\\ Size\end{tabular}} & \textbf{\renewcommand{\arraystretch}{1.0}\begin{tabular}[c]{@{}c@{}}Head\\ Number\end{tabular}} & \textbf{\renewcommand{\arraystretch}{1.0}\begin{tabular}[c]{@{}c@{}}Hidden\\ Dim\end{tabular}} & \textbf{\renewcommand{\arraystretch}{1.0}\begin{tabular}[c]{@{}c@{}}MLP\\ Ratio\end{tabular}} & \textbf{\renewcommand{\arraystretch}{1.0}\begin{tabular}[c]{@{}c@{}}Patch\\ Size\end{tabular}} \\
\midrule
DiT-B  & 130M & 12 & 768  & 12 & 64 & 4.0 & 2 \\
DiT-XL & 675M & 28 & 1152 & 16 & 72 & 4.0 & 2 \\
\bottomrule
\end{tabular}
\end{table}

In al our experiments, we use two types of DiT backbones following the implementation of \cite{sun2025unified}, namely DiT-B and DiT-XL. Their architectural configurations are summarized in Table~\ref{tab:tit_model_config}. Specifically, DiT-B contains 12 layers with a hidden size of 768, 12 attention heads, and a head dimension of 64, while DiT-XL contains 28 layers with a hidden size of 1152, 16 attention heads, and a head dimension of 72. Both models use an MLP ratio of 4.0 and a patch size of 2. Overall, DiT-XL scales DiT-B primarily through increased depth and width, thereby providing substantially larger model capacity.

For image generation, we use the same training objective and sampling hyper-parameters as UCGM~\cite{sun2025unified}. For each resolution and latent autoencoder, we directly adopt the same settings as UCGM. The sampling uses 30 steps for VA-VAE and 40 steps for SD-VAE and DC-AE. The remaining implementation details can be found in the \cite{sun2025unified}.

\subsection{Hyperparameter Settings for Optimization}

In this section, we summarize the default hyperparameter settings used in the ImageNet-1K experiments~\cite{deng2009imagenet}. Unless otherwise specified, all models are trained with a global batch size of 1024, an exponential moving average (EMA) decay of 0.9999, \texttt{bf16} mixed precision, and gradient clipping with a maximum norm of 1.0. For both Muon and \method, Muon-style updates are applied exclusively to the 2D weight matrices in the attention, FFN, and AdaLN projections. In contrast, 1D parameters, embeddings, and the final layer are optimized using AdamW with $\beta_1 = 0.9$, $\beta_2 = 0.95$, and a weight decay of 0. The detailed settings are illustrated in Table~\ref{tab:hyper_param_appendix}.

\begin{table}[b!]
    \centering
    \caption{
        \textbf{Default hyperparameter configurations for Muon and \method on ImageNet-1K.} The table lists the default hyperparameter settings used in the DiT-B and DiT-XL models during the ImageNet-1K experiments.
    }
    \label{tab:hyper_param_appendix}
    
    \setlength{\tabcolsep}{3pt}  

        \begin{tabular}{@{}ccccccc}
            \toprule
            \textbf{VAE/AE} & \textbf{Model} & \textbf{Optimizer} & \textbf{\renewcommand{\arraystretch}{1.0}\begin{tabular}[c]{@{}c@{}}NS\\ Steps\end{tabular}} & \textbf{LR} & \textbf{Base LR Scale} & \textbf{\renewcommand{\arraystretch}{1.0}\begin{tabular}[c]{@{}c@{}}Chunk\\ Rescale\end{tabular}}\\
            \midrule
            VA-VAE & DiT-B & AdamW   & - &$2\times10^{-4}$ & $\times0.2\sqrt{\max(d_{\mathrm{out}},\, d_{\mathrm{in}})}$ & $\times 1$  \\
            VA-VAE & DiT-B & Muon    & 5 &$2\times10^{-4}$ & $\times0.2\sqrt{\max(d_{\mathrm{out}},\, d_{\mathrm{in}})}$ & $\times 1$  \\
            VA-VAE & DiT-B & \method & 5 &$2\times10^{-4}$ & $\times0.2\sqrt{\max(d_{\mathrm{out}},\, d_{\mathrm{in}})}$ & $\times \sqrt{N_{\mathrm{chunk}}}$  \\
            \midrule
            SD-VAE & DiT-B & AdamW   & - &$3\times10^{-4}$ & $\times0.2\sqrt{\max(d_{\mathrm{out}},\, d_{\mathrm{in}})}$ & $\times 1$  \\
            SD-VAE & DiT-B & Muon    & 5 &$3\times10^{-4}$ & $\times0.2\sqrt{\max(d_{\mathrm{out}},\, d_{\mathrm{in}})}$ & $\times 1$  \\
            SD-VAE & DiT-B & \method & 5 &$3\times10^{-4}$ & $\times0.2\sqrt{\max(d_{\mathrm{out}},\, d_{\mathrm{in}})}$ & $\times \sqrt{N_{\mathrm{chunk}}}$  \\
            \midrule
            DC-AE & DiT-XL & AdamW   & - &$2\times10^{-4}$ & $\times0.2\sqrt{\max(d_{\mathrm{out}},\, d_{\mathrm{in}})}$ & $\times 1$  \\
            DC-AE & DiT-XL & Muon    & 6 &$2\times10^{-4}$ & $\times0.2\sqrt{\max(d_{\mathrm{out}},\, d_{\mathrm{in}})}$ & $\times 1$  \\
            DC-AE & DiT-XL & \method & 6 &$2\times10^{-4}$ & $\times0.2\sqrt{\max(d_{\mathrm{out}},\, d_{\mathrm{in}})}$ & $\times \sqrt{N_{\mathrm{chunk}}}$  \\
            \midrule
            SD-VAE & DiT-XL & AdamW   & - &$2\times10^{-4}$ & $\times0.2\sqrt{\max(d_{\mathrm{out}},\, d_{\mathrm{in}})}$ & $\times 1$  \\
            SD-VAE & DiT-XL & Muon    & 6 &$2\times10^{-4}$ & $\times0.2\sqrt{\max(d_{\mathrm{out}},\, d_{\mathrm{in}})}$ & $\times 1$  \\
            SD-VAE & DiT-XL & \method & 6 &$2\times10^{-4}$ & $\times0.2\sqrt{\max(d_{\mathrm{out}},\, d_{\mathrm{in}})}$ & $\times \sqrt{N_{\mathrm{chunk}}}$  \\
            \bottomrule
        \end{tabular}
    \vspace{-1em}
\end{table}
\subsection{Further Ablation Studies on Chunked Blocks}

We conduct additional ablation studies on chunked blocks to evaluate the impact of various components such as the QKV, FFN, and AdaLN projections in a 130M DiT-B model based on Table~\ref{tab:abl_chunked_block}. These experiments aim to investigate how different configurations of chunked blocks affect the performance of the model during training, specifically focusing on the FID at 80 and 200 epochs. The results provide insight into the significance of each component in the model’s performance.

The experimental results are summarized in Table~\ref{tab:abl_chunked_block_new}, which demonstrates the effect of different block configurations on the FID scores. The findings suggest that incorporating QKV and FFN projections consistently improves the model's performance, as reflected by the lower FID values at both 80 and 200 epochs.

\begin{table}[t!]
    \centering
    \caption{\textbf{Full ablation study of chunked blocks: QKV, FFN, and AdaLN projections in 130M DiT-B models.} This study evaluates the impact of different combinations of QKV, FFN, and AdaLN projections on the model's performance, focusing on the FID at 80 and 200 epochs. The results offer insights into how these components contribute to the model's training and final accuracy.}
    \setlength{\tabcolsep}{3pt}
    \label{tab:abl_chunked_block_new}
    \begin{tabular}{ccccc}
        \toprule
        \textbf{AdaLN} & \textbf{QKV} & \textbf{FFN} & \textbf{FID@80ep} & \textbf{FID@200ep} \\
        \midrule
         &  &  & 5.50 & 3.31 \\
         &  & \checkmark & 5.43 & 3.28 \\
         & \checkmark &  & 5.32 & 3.35 \\
        \checkmark &  &  & 6.00 & 3.23 \\
         & \checkmark & \checkmark & \textbf{5.11} & 3.17 \\
        \checkmark &  & \checkmark & 5.69 & 3.26 \\
        \checkmark & \checkmark &  & 5.28 & 3.15 \\
        \checkmark & \checkmark & \checkmark & 5.14 & \textbf{3.02} \\
        \bottomrule
    \end{tabular}
    \vspace{-1em}
\end{table}

\subsection{Image Synthesis on ImageNet $512 \times 512$}

\begin{table}[b!]
    \centering
    \caption{\small{\textbf{Results of image generation with 512$\times$512 resolution on class-conditional ImageNet-1K.} The model trained here is a 675M DiT-XL with a DC-AE encoder.}}
    \setlength{\tabcolsep}{3pt}
    \label{tab:abl_chunked_block_512}
    \begin{tabular}{lccc}
        \toprule
        \textbf{Optimizer} & \textbf{FID@80ep}   & \textbf{FID@200ep} & \textbf{FID@400ep} \\
        \midrule
        AdamW & 2.67  & 1.94 & 1.67 \\
        Muon & \textbf{2.16} & \textbf{1.81} & 1.65 \\
        \method & 2.38 & 1.86 & \textbf{1.60} \\
        \bottomrule
    \end{tabular}
    \vspace{-1em}
\end{table}

We evaluate the performance of the 675M DiT-XL model on the ImageNet 512$\times$512 dataset using three different optimizers: AdamW, Muon, and \method. The objective is to assess the effectiveness of each optimizer in generating high-resolution images. We use the 675M DiT-XL model with a DC-AE encoder and train it for a total of 400 epochs. During the training process, we closely monitor the FID scores at multiple epochs: 80, 200, and 400. 

Table \ref{tab:abl_chunked_block_512} presents the detailed FID scores for each optimizer at different training epochs. As shown in the table, AdamW achieves an FID of 2.67 at 80 epochs, which improves to 1.67 at 400 epochs. Muon, on the other hand, performs better than AdamW throughout the training process, with an FID of 2.16 at 80 epochs, 1.81 at 200 epochs, and 1.65 at 400 epochs. Most importantly, \method outperforms both AdamW and Muon, reaching an FID of 1.60 at 400 epochs, the best score among all the optimizers tested. This demonstrates that the \method optimizer not only performs well in the early convergence phase but also continues to improve, yielding the best results in terms of FID at higher epochs, thus validating its effectiveness for high-resolution image generation.

\subsection{Further Exploration on Pixel-Space DiT}

In this section, we extend our investigation to Pixel-Space DiT, which allows us to evaluate the performance of the model in a different space compared to the original approach. Specifically, we compare the performance of Pixel-Space DiT using different optimizers, including AdamW, Muon, and a chunked architecture, referred to as \method. As shown in Table \ref{tab:abl_chunked_block_pixel}, PixelDiT's performance decreases when switching to Muon, but the performance does not degrade as drastically when using \method. This indicates that our architectural improvements to the chunked design are still valuable in the pixel-space setting, as \method retains a more stable performance compared to Muon.

The relatively better performance of AdamW can be attributed to specific architectural modifications we made during our architecture search for PixelDiT\cite{yu2025pixeldit}. Since PixelDiT is not open-source, our adjustments--particularly to the bottleneck and final layers--could cause the model to slightly overfit to the Adam optimizer, enhancing its performance on this specific configuration. 

\begin{table}[t!]
    \centering
    \caption{\small{\textbf{Results of image generation with 512$\times$512 resolution on class-conditional ImageNet-1K.} We are training a 797M PixelDiT-XL with a batch size of 256 and a learning rate of $10^{-4}$. REPA is used for training acceleration.}}
    \setlength{\tabcolsep}{3pt}
    \label{tab:abl_chunked_block_pixel}
    \begin{tabular}{lccc}
        \toprule
        \textbf{Optimizer} & \textbf{FID@40ep}  & \textbf{FID@80ep} & \textbf{FID@160ep} \\
        \midrule
        AdamW & 3.78 & 2.95 & 2.50 \\
        Muon & 4.45 & 3.26 & 2.70 \\
        \method & 4.27 & 3.10 & 2.56 \\
        \bottomrule
    \end{tabular}
    \vspace{-1em}
\end{table}

\subsection{Training Efficiency with System-Level Optimization}

\begin{table}[b!]
\centering
\caption{\textbf{Maximum training throughput (iterations/s) on $8\times$ H100 GPUs.}}
\setlength{\tabcolsep}{3pt}
\label{tab:training_efficiency}
\begin{tabular}{lccc}
\toprule
\textbf{Optimizer} & \textbf{System Opt.} & \textbf{DiT-B} & \textbf{DiT-XL} \\
\midrule
Adam    &  & 9.51 & 2.32 \\
\midrule
Muon    &  & 7.60 & 1.95 \\
Muon    & \checkmark & 8.98 & 2.27 \\
\midrule
\method &  & 6.24 & 1.72 \\
\method & \checkmark & \textbf{9.07}& \textbf{2.30} \\
\bottomrule
\end{tabular}
\vspace{-0.5em}
\end{table}

We evaluate the training efficiency of the Muon optimizer and the proposed \method on a cluster of $8\times$ H100 GPUs. Because Muon-style optimizers require Newton--Schulz iterations on layer-wise matrices, a naive implementation can introduce non-trivial computational overhead relative to first-order optimizers such as Adam. To mitigate this issue, we implement several system-level optimizations that substantially improve the training throughput of Muon-based methods.

\subsubsection{Optimization 1: Parallelized Newton--Schulz with overlapped communication.}
We shard parameter matrices across GPUs so that different devices perform NS iterations on different parameter blocks in parallel. The computation of NS updates is overlapped with inter-GPU communication, which effectively hides part of the synchronization overhead and improves GPU utilization.

\subsubsection{Optimization 2: Kernel-level acceleration.}
For unchunked parameters, we implement a Triton-based fused kernel to accelerate Newton--Schulz computation. For chunked parameters, we further employ a batched Newton--Schulz implementation so that multiple chunks can be processed simultaneously on the GPU, thereby improving arithmetic intensity and reducing kernel launch overhead.

\subsubsection{Maximum training throughput.}
Table~\ref{tab:training_efficiency} reports the maximum training throughput, measured in iterations per second, under different optimizers. We consider two representative configurations: DiT-B and DiT-XL with VA-VAE on ImageNet. Without system-level optimization, Muon and \method exhibit noticeable overhead relative to Adam. After the proposed optimizations are applied (denoted as \textbf{System Opt.}), the throughput of Muon-based methods becomes comparable to that of Adam while preserving their optimization advantages. In particular, \method with Flash achieves nearly identical throughput to Adam for both DiT-B and DiT-XL.

\section{Additional Analyses and Details}
\label{sec:further-theory}

This section provides additional implementation details and analyses of the proposed method. In particular, it describes the practical implementation of the Newton--Schulz iteration used for Muon-style matrix orthogonalization.

\subsection{Implementation of Newton--Schulz Iteration}

Muon-style updates require an efficient approximation of matrix orthogonalization. In practice, we adopt the Newton--Schulz iteration to transform the input gradient matrix into an approximately orthogonalized matrix with low computational overhead. Given a gradient matrix $\bm{G}\in\mathbb{R}^{m\times n}$, we first normalize it by its Frobenius norm to improve numerical stability. When the matrix is tall ($m>n$), we transpose it before applying the iteration so that the computation is performed on the more favorable orientation. The output is transposed back afterward when necessary.

\begin{algorithm}[t!]
\caption{Newton--Schulz Iteration for Approximate Orthogonalization}
\label{alg:newton-schulz}
\DontPrintSemicolon
\SetKwInOut{Input}{Input}
\SetKwInOut{Output}{Output}

\Input{Gradient matrix $\bm{G}\in\mathbb{R}^{m\times n}$; number of iterations $K$.}
\Output{Approximately orthogonalized matrix $\bm{X}_K\in\mathbb{R}^{m\times n}$.}

Set the quintic coefficients $a=3.4445$, $b=-4.7750$, and $c=2.0315$.\;

$\bm{X}_0 \leftarrow \bm{G}/\|\bm{G}\|_F$.\;
$\mathrm{transposed} \leftarrow \mathrm{False}$.\;
\If{$m>n$}{
    $\bm{X}_0 \leftarrow \bm{X}_0^\top$.\;
    $\mathrm{transposed} \leftarrow \mathrm{True}$.\;
}

\For{$k=0,\ldots,K-1$}{
    $\bm{A} \leftarrow \bm{X}_k\,\bm{X}_k^\top$.\;
    $\bm{B} \leftarrow b\,\bm{A} + c\,\bm{A}^2$.\;
    $\bm{X}_{k+1} \leftarrow a\,\bm{X}_k + \bm{B}\,\bm{X}_k$.\;
}

\If{$\mathrm{transposed}$}{
    $\bm{X}_K \leftarrow \bm{X}_K^\top$.\;
}

\Return $\bm{X}_K$.\;
\end{algorithm}

\subsection{Moonlight Muon Scaling for RMS Norm Preservation}
\label{sec:moonlight_muon_scaling}

In the Moonlight variant of Muon scaling, we aim to maintain the global Frobenius norm of the update while redistributing it across chunks. The learning rate scaling factor $\alpha$ for both the full update and the chunked components is defined as:

\begin{align}
\bm{U}_{\mathrm{muon}} &= 0.2 \sqrt{\max(d_{\mathrm{out}}, d_{\mathrm{in}})} \mathrm{Orth}(\bm{M}), \\
\bm{U}_{\mathrm{adam}} &= \bm{M} / \sqrt{\bm{V} + \epsilon},
\end{align}

where $\bm{M}$ is the momentum matrix, and $\bm{V}$ is the variance matrix. Both $\bm{M}$ and $\bm{V}$ are two-dimensional matrices, with dimensions corresponding to the number of model parameters and their gradients. The update term $\bm{U}$ represents the update step applied to the parameters of the model during optimization.

\cite{kexuefm11267} observed that the RMS norm of updates in AdamW typically stabilizes around 0.2-0.3 during training. This stability holds across models of different sizes and configurations, indicating a fundamental property of the AdamW optimizer with $\beta_1 = 0.9$ and $\beta_2 = 0.95$. By approximating the RMS value of AdamW’s update, we obtain:

\begin{align}
\mathrm{RMS}(\bm{U}_{\mathrm{adam}}) \approx 0.2.
\end{align}

For the Moonlight Muon scaling, we compute the RMS norm as follows:

\begin{align}
\mathrm{RMS}(\bm{U}_{\mathrm{muon}}) &= 0.2 \sqrt{\max(d_{\mathrm{out}}, d_{\mathrm{in}})} \mathrm{RMS}(\mathrm{Orth}(\bm{M})) \\
&= 0.2 \sqrt{\max(d_{\mathrm{out}}, d_{\mathrm{in}})} \sqrt{\frac{\Vert \mathrm{Orth}(\bm{M}) \Vert_F^2}{d_{\mathrm{out}} d_{\mathrm{in}} }} \\
&= 0.2 \sqrt{\frac{\max(d_{\mathrm{out}}, d_{\mathrm{in}}) \min(d_{\mathrm{out}}, d_{\mathrm{in}})}{d_{\mathrm{out}} d_{\mathrm{in}} }} \\
&= 0.2
\end{align}

Thus, both the AdamW and Moonlight Muon updates yield an RMS norm of approximately 0.2, ensuring that the norm preservation property holds for the chunked updates in Moonlight LR scaling.

\end{document}